\documentclass[twocolumn,
]{ceurart}

\usepackage{enumitem}

\begin{document}

\copyrightyear{2024}
\copyrightclause{Copyright for this paper by its authors.
  Use permitted under Creative Commons License Attribution 4.0
  International (CC BY 4.0).}

\conference{SEPLN-CEDI2024: Seminar of the Spanish Society for Natural Language Processing at the 7\textsuperscript{th} Spanish Conference on Informatics}

\title{Dancing in the syntax forest: fast, accurate and explainable sentiment analysis with SALSA}

\author[1]{Carlos Gómez-Rodríguez}[%
orcid=0000-0003-0752-8812,
email=carlos.gomez@udc.es,
url=https://www.grupolys.org/~cgomezr,
]
\address[1]{Universidade da Coruña, CITIC. Department of Computer Science and Information Technologies. 15071 A Coruña, Spain}

\address[2]{Universidade da Coruña, Technology Transfer Unit (OTRI). 15071 A Coruña, Spain}

\author[1]{Muhammad Imran}[%
orcid=0000-0002-4124-7929,
email=m.imran@udc.es,
url={https://scholar.google.com/citations?user=cLXCYOcAAAAJ},
]

\author[1]{David Vilares}[%
orcid=0000-0002-1295-3840,
email=david.vilares@udc.es,
url=https://www.grupolys.org/~david.vilares/,
]

\author[2]{Elena Solera}[%
orcid=0000-0003-3541-8303,
email=elena.solera@udc.es,
]

\author[1]{Olga Kellert}[%
orcid=0000-0001-8601-8305,
email=o.kellert@udc.es,
url={https://scholar.google.com/citations?user=XLj-Ol4AAAAJ},
]

\begin{abstract}
  Sentiment analysis is a key technology for companies and institutions to gauge public opinion on products, services or events. However, for large-scale sentiment analysis to be accessible to entities with modest computational resources, it needs to be performed in a resource-efficient way. While some efficient sentiment analysis systems exist, they tend to apply shallow heuristics, which do not take into account syntactic phenomena that can radically change sentiment. Conversely, alternatives that take syntax into account are computationally expensive. The SALSA project, funded by the European Research Council under a Proof-of-Concept Grant, aims to leverage recently-developed fast syntactic parsing techniques to build sentiment analysis systems that are lightweight and efficient, while still providing accuracy and explainability through the explicit use of syntax. We intend our approaches to be the backbone of a working product of interest for SMEs to use in production. 
\end{abstract}


\maketitle

\section{Introduction}

We describe the project ``efficient Syntactic Analysis for Large-scale Sentiment Analysis (SALSA)'', which is currently being developed by members of the LyS research group at the CITIC research center (Universidade da Coruña). The project is funded by the European Research Council (ERC) with a budget of 150 000 €, under the Proof-of-Concept (PoC) grant scheme, with grant agreement number 101100615. The Proof-of-Concept grant program intends to bridge the gap between basic research results obtained in ERC projects and the early phases of their commercialization, by exploring the innovation potential of the work and bringing it closer to market. The SALSA project started on February 2023 and is scheduled to run until July 2024, with the possibility of an extension.

Given the nature of PoC grants, the project has a scientific research component as well as an innovation and technology transfer component. In Sections~\ref{sec:motivation} and~\ref{sec:methodology}, we describe the motivation and methodology, respectively, focusing on the scientific part of the project. Section~\ref{sec:planning} describes the planning and the project team, with a more global focus that includes the transfer parts as well.

\section{Motivation}
\label{sec:motivation}

\paragraph{The problem.} The Internet and social media are major platforms where people express their views and share experiences about a variety of topics, including products, services, and events. This results in a wealth of information that can be leveraged to understand public perception, pinpoint product strengths and weaknesses, discover and address people's needs, or track political and market trends. This is the goal of \emph{opinion mining} or \emph{sentiment analysis} systems, which can analyze text and extract sentiment information from it. For instance, a company that releases a new smartphone could gather user opinions from social networks by collecting messages mentioning the product. Given each such message, a sentiment analysis system can determine whether it contains a positive, negative or neutral opinion, both towards the phone as a whole or towards specific aspects like camera or battery.

While many approaches to sentiment analysis have been proposed, an extant challenge lies in the absence of systems that can efficiently process opinions while considering the complex structure of language. While efficient sentiment analysis systems exist, like SentiStrength~\cite{sentistrength}, they rely on shallow heuristic methods that count sentiment words but may struggle with sentences where the overall sentiment is shaped by the grammatical structure. For example, SentiStrength produces the exact same result for ``This phone is expensive, and it's not at all good'' and ``This phone is good, and it's not at all expensive'': both sentences actually have opposite sentiment polarities, but the heuristics cannot correctly capture the scope of negation and its influence on how the sentiment should be interpreted. 

An alternative is to use approaches that take syntactic stucture into account. Syntax can be injected into a sentiment analysis model either explicitly, by using a parser~\cite{VilAloGomNLE2015,VilGomAloKBS2017}, or implicitly, by using pretrained language models~\cite{hoang-etal-2019-aspect} which have been shown to internally represent syntax~\cite{VilStrSogGomAAAI2020}. 
The advantage of the former approach over the latter is that it can provide explainability, as an explicit parse tree can be used to show how the polarity is obtained, while the latter can be better in terms of raw accuracy.
However, both of these approaches have a high computational cost: pretrained models (especially newer large language models) have considerable CPU and memory requirements~\citep{strubell-etal-2019-energy}, while syntactic parsers have traditionally been slow~\cite{GomSEPLN2017,GomAloVilAIRE2019}. This makes existing syntactically-guided sentiment analysis approaches infeasible for small entities that have a constrained computational budget.

\paragraph{The solution.} The SALSA project proposes to leverage a breakthrough achieved by the previous ERC-funded FASTPARSE project~\cite{GomSEPLN2017} to build sentiment analysis systems that are efficient and, at the same time, use explicit syntax to improve accuracy and provide explainability. The FASTPARSE project had the goal to substantially improve the speed of syntactic parsing, and was a resounding success, advancing the state of the art in both parsing speed and accuracy several times, and achieving accurate parsers that improved speed by an order or magnitude over previous approaches.

In particular, within the FASTPARSE project, we defined a new syntactic parsing paradigm that casts parsing as a sequence labeling task by defining encodings that can represent syntax trees by means of discrete labels associated with words. This was shown to work for the two main families of grammatical formalisms, with encodings that proved viable for constituency parsing~\cite{GomVilEMNLP2018} and dependency parsing~\cite{StrVilGomNAACL2019},
and new encodings have kept being defined since then that further push parsing accuracy~\cite{amini-etal-2023-hexatagging,GomRocVilEMNLP2023}.

For the purposes of this project, the approach to parsing as sequence labeling has two especially relevant advantages:
\begin{enumerate}
\item Pluggability: parse trees are represented with tags associated with words, a standard representation that is easy to feed to downstream tasks, be it as features~\cite{wang-etal-2019-best} or by using multitask learning~\cite{StrVilGomACL2019}.
\item Speed: these parsers can parse around a thousand sentences per second on a standard consumer GPU~\cite{GomVilEMNLP2018,StrVilGomNAACL2019,anderson-gomez-rodriguez-2021-modest}.
\end{enumerate}
Our proposal is, thus, to use this new parsing paradigm to create efficient syntax-guided sentiment analysis systems that can operate at a large scale with modest computational resources.
This will democratize accurate large-scale sentiment analysis technology, putting it within reach of smaller institutions and companies that cannot afford to deploy slow parsers or language models.

\section{Methodology}
\label{sec:methodology}

Our project addresses two variants (or subtasks) of sentiment analysis:
\begin{enumerate}[label=(\alph*),leftmargin=3\parindent]
\item Polarity classification: a coarse-grained task where the input is a text, such as a social network message, and the output is its polarity (i.e. a representation, typically in a discrete scale, of how positive or negative the global opinion expressed in the text is).
\item Fine-grained sentiment analysis, often called aspect-based sentiment analysis~\cite{brauwers}: in this task, the goal is not to obtain a single value for the whole text, but individual opinions about the different targets that may appear in the message (for example, the text ``I really like this phone's camera, but the battery life is not acceptable'' contains both a highly positive opinion about the camera and a negative one about the battery).
\end{enumerate}

For this purpose, we will explore three approaches to integrate fast syntactic parsing into solving these two sentiment analysis tasks:
\begin{itemize}
\item Rule-based pipeline approach: we start by using a fast dependency parser to obtain the syntactic tree for an input sentence. We then use syntax-based rules and a polarity lexicon to navigate the tree, assigning polarity values to its syntactic components (addressing subtask (b) above). These values can then be integrated by the rules to obtain an overall polarity for the whole sentence (addressing subtask (a)). The syntax-based rules are specifically designed to address syntactic phenomena that influence opinion polarity, such as negation (for instance, in the example above, "not acceptable" indicates a negative sentiment because the negation alters the positive meaning of "acceptable"), intensification (the phrase "really like" signifies a stronger preference than merely "like"), and adversatives (where "but" juxtaposes a positive statement against a negative one). It is worth noting that this kind of approach have been applied in the past, particularly for subtask (a) (see~\citep{VilAloGomNLE2015,VilGomAloKBS2017}), but the use of slow parsers at the time rendered large-scale application costly.
\item Multitask learning approach: thanks to the parsing-as-sequence-labeling approach, we can train a multitask learning model to perform sentiment analysis and syntactic parsing simultaneously. This allows each task to benefit from the insights of the other implicitly, without relying on explicit syntax-based rules. For subtask (a), this technique can be implemented by formulating the sentiment analysis sequence labeling task in such a way that the polarity of the whole sentence is represented by a tag associated with the last word. For subtask (b), we will design a sequence labeling encoding to represent fine-grained opinoins.
\item Integrated approach: this approach is specific to subtask (b). We represent the sentiment structure of the sentence as a tree. This means that we can apply the same algorithms that we use for syntactic parsing directly to perform sentiment analysis. Furthermore, this strategy can be combined with the multitask learning approach described above, enabling the training of a sequence labeling model to simultaneously produce two types of trees: one for syntactic parsing and another for sentiment analysis.
\end{itemize}

Each of the described approaches will be evaluated on freely-available sentiment analysis corpora. We will take a multilingual focus, ensuring that our systems work on a diverse range of languages (for the dependency parsing part, this is ensured by the use of Universal Dependencies~\cite{ud212}). For the initial value proposition, we will focus on Spanish and English, since our main target is the local Spanish market. However, the core technology should be available to easily adapt to more languages when needed.

In particular, a good starting point when it comes to datasets are the SemEval 2022 Task 10 corpora~\citep{barnes-etal-2022-semeval}, as they cover five different languages and their annotation includes individual polarity targets, thus supporting both our subtasks (a) and (b). Given our focus on the Spanish market, we are also working with a corpus of TripAdvisor reviews in Spanish, from Rest-Mex 2023~\cite{restmex}. This also allows us to evaluate performance in a noisy context. Some preliminary results of the first (rule-based pipeline) approach on the latter corpus can be found in~\cite{KelZamMatGomIberLEF2023}. 

\section{Planning and Team}
\label{sec:planning}

The SALSA project is organized into three work packages. The first one addresses the scientific research part of the project, and hence it is the one we have addressed in Sections 2 and 3. The other two correspond to the innovation and technology transfer parts of the project. In particular, the work packages are as follows:

\begin{itemize}
\item WP1: Technical and Performance Validation. This involves three tasks:
\begin{itemize}
\item Validation of fast parsers for polarity classification: the main objective is to adapt the parsers developed in the FASTPARSE project to work on polarity classification tasks.
To achieve this, we have created linguistic rules that, when integrated with a parser's output, allow us to determine the polarity of sentences by analyzing the words and syntactic structures in the text~\cite{KelZamMatGomIberLEF2023}. Additionally, we are exploring a purely machine learning-based method that does not rely on these rules. This method involves training a parser capable of sequence labeling in conjunction with a sequence model that uses discrete tags to represent sentence polarity, employing a multitask learning framework. The success of this task is assessed through the validation of various models' ability to classify polarity, as evidenced by performance metrics.
\item Validation of fast parsers for fine-grained sentiment analysis: in this task, we aim to go beyond polarity classification by employing syntactic parsers for more detailed sentiment analysis. The goal is to analyze sentiments at a more granular level, capturing specific opinions within a sentence. This involves recognizing how a single sentence might simultaneously provide positive feedback on one aspect of a product while criticizing another. To achieve this, we will adapt the approaches based on linguistic rules and on multitask learning to process fine-grained information: the rules will be used to infer sentiment information for smaller linguistic units apart from the whole sentence, and a sequence labeling encoding will be used to represent fine-grained sentiment information for the multitask setup. In addition, we will try a third setup where the fine-grained sentiment information will be represented as a tree, so that syntactic parsers can learn to output it directly. 
\item Cross-domain generalization. The performance of sentiment analysis models trained on a given domain (e.g. restaurant reviews) often degrades when they are used in a different domain (e.g. movie reviews). Since our project is targeted primarily towards making the technology accessible to small entities, which will often lack access to high-quality in-domain corpora, it is important to ensure that our systems can adapt to different domains. In this respect, rule-based approaches have been shown to be better than purely supervised approaches~\cite{VilGomAloKBS2017}. In this task, we will evaluate the generalization capabilities of the various models developed in the previous two tasks to different domains, and explore specific domain adaptation approaches (like the adaptation of sentiment dictionaries~\cite{VilAloGomNLE2015}) if needed.
\end{itemize}
\item WP2: Market Research and Validation. This involves three tasks:
\begin{itemize}
\item Market Analysis. We have performed a study to identify current sentiment analysis industry trends, market need and market size at the European level. This study, which was subcontracted to an external consultancy company, includes the potential market niches for our sentiment analysis models, identification of comparative technologies and information on potential user needs.
\item Validation of value proposition. This involves conducting interviews, as well as Design Thinking workshops, to identify target user needs and validate our proposition.
\item Commercial viability: product-market fit. This task consists in involving potential users (identified in the previous tasks) in iterative testing and validation of the developed models.
\end{itemize}
\item WP3: Pre-Technology Transfer. This includes two tasks:
\begin{itemize}
\item Creating SALSA business model canvas. We will refine our business model canvas through a process of continuous improvement. This will involve defining, enhancing, and validating our value proposition, primary revenue streams, and cost structures, alongside identifying the crucial partners, customers, channels, and business relationships that form the core components of an open source business model. We will assess the viability of establishing a spin-off to offer service agreements related to SALSA.
\item Building a business environment. The aim of this task is to establish contact with potential partners and stakeholders that can advance the TRL and scale the technology, as well as with potential future investors. We will also considering the possibility of preparing an EIC Transition proposal, depending on the outcomes of previous tasks.
\end{itemize}
\end{itemize}

The project team is composed by the authors of this paper: a postdoctoral researcher (O. Kellert), an MSCA predoctoral researcher (M. Imran), a technical manager (D. Vilares) and a project manager (E. Solera), led by the PI (C. Gómez-Rodríguez). In addition, some of the market-oriented tasks of the project are being carried out with the help of Matical Innovation, which has been subcontracted for that purpose.

\begin{acknowledgments}
This project has received funding by the European Research Council (ERC), under the
Horizon Europe research and innovation programme (SALSA, grant agreement No 101100615).
\end{acknowledgments}

\bibliography{sample-ceur}

\appendix

\end{document}